\documentclass[11pt]{article}
\usepackage[utf8]{inputenc}
\usepackage{cmap}
\usepackage[T1]{fontenc}
\usepackage[margin=1in]{geometry}
\usepackage{amsmath,amssymb}
\usepackage{graphicx}
\usepackage{booktabs}
\usepackage[round]{natbib}
\usepackage{newunicodechar}
\usepackage{xcolor}
\usepackage[colorlinks=true,linkcolor=black,citecolor=blue,urlcolor=blue]{hyperref}
\usepackage{caption}
\usepackage{longtable}
\usepackage{calc}

\usepackage{array}

\newunicodechar{‖}{\ensuremath{\|}}
\newunicodechar{α}{\ensuremath{\alpha}}
\newunicodechar{β}{\ensuremath{\beta}}
\newunicodechar{ρ}{\ensuremath{\rho}}
\newunicodechar{σ}{\ensuremath{\sigma}}
\newunicodechar{Σ}{\ensuremath{\Sigma}}
\newunicodechar{Φ}{\ensuremath{\Phi}}
\newunicodechar{φ}{\ensuremath{\varphi}}
\newunicodechar{ℓ}{\ensuremath{\ell}}
\newunicodechar{η}{\ensuremath{\eta}}
\newunicodechar{λ}{\ensuremath{\lambda}}
\newunicodechar{Δ}{\ensuremath{\Delta}}
\newunicodechar{χ}{\ensuremath{\chi}}
\newunicodechar{ε}{\ensuremath{\epsilon}}
\newunicodechar{μ}{\ensuremath{\mu}}
\newunicodechar{×}{\ensuremath{\times}}
\newunicodechar{→}{\ensuremath{\rightarrow}}
\newunicodechar{≤}{\ensuremath{\leq}}
\newunicodechar{≥}{\ensuremath{\geq}}
\newunicodechar{∝}{\ensuremath{\propto}}
\newunicodechar{≈}{\ensuremath{\approx}}
\newunicodechar{∈}{\ensuremath{\in}}
\newunicodechar{²}{\ensuremath{^{2}}}
\newunicodechar{₀}{\ensuremath{_{0}}}
\newunicodechar{₁}{\ensuremath{_{1}}}
\newunicodechar{₂}{\ensuremath{_{2}}}
\newunicodechar{₃}{\ensuremath{_{3}}}
\newunicodechar{₄}{\ensuremath{_{4}}}
\newunicodechar{₅}{\ensuremath{_{5}}}
\newunicodechar{ᵢ}{\ensuremath{_{i}}}
\newunicodechar{·}{\ensuremath{\cdot}}
\newunicodechar{−}{\ensuremath{-}}
\newunicodechar{∼}{\ensuremath{\sim}}
\newunicodechar{Ž}{\v{Z}}
\newunicodechar{ž}{\v{z}}
\newunicodechar{ć}{\'c}
\newunicodechar{č}{\v{c}}
\newunicodechar{é}{\'e}
\newunicodechar{ï}{\"i}
\newunicodechar{“}{``}
\newunicodechar{”}{''}
\newunicodechar{’}{'}
\newunicodechar{‘}{`}
\newunicodechar{—}{---}
\newunicodechar{–}{--}

\title{\textbf{What Does the Weight Norm Control in Grokking?\\
Logit-Scale Mediation under Cross-Entropy}}
\author{Truong Xuan Khanh\\ H\&K Research Studio, Clevix LLC\\ Hanoi, Vietnam\\ \texttt{khanh@clevix.vn}}
\date{}

\input{glyphtounicode}
\begin{document}
\maketitle
\begin{abstract}
Grokking is generalization that appears long after a network has fit its training data. The grokking
delay grows with the weight norm, and holding the norm fixed with a clamp reproduces a clean dose
response, which has invited the reading that the weight norm sets the timescale. We show that this
reading is incomplete. The clamp holds the scalar norm fixed by rescaling the weights, but rescaling also
raises the logit scale, so the norm and the logit scale move together and the clamp does not separate
them. We separate them with a non-trainable output temperature that divides the logits before the loss
and is not part of the norm. Holding the total weight norm fixed and varying only this temperature slides
the delay across the full range produced by raising the norm, and matching the effective logit scale back
to its baseline value recovers $0.83$ (95\% CI $[0.82,0.85]$) and $0.89$ ($[0.88,0.91]$) of the
norm-induced delay at two moduli. Across a grid of norms and temperatures the delay collapses onto the
effective logit scale, which alone explains $97\%$ of its variance while the norm dose adds $1$--$2\%$.
Under cross-entropy, then, the weight-norm effect on the delay runs
through the logit scale and the softmax saturation it causes, not through the scalar norm itself. The
same intervention does nothing of the sort under mean-squared error: there the effective logit scale is
pinned near one and cannot be moved, the norm effect is about half the size, and it is carried by a
separate route. The mechanism is therefore loss-dependent. A control shows the temperature acts on the
delayed-generalization phase and not on memorization time, ruling out a gradient-magnitude artifact, an
independent float64 audit reaches the same softmax-saturation channel by precision rather than by
intervention, and a no-LayerNorm transformer reproduces the effect. Forking arms from one identical state
shows the delay tracks the held norm value, not the rescaling operation, closing the clamp-artifact
objection. Claims are confined to the $\ell_2$/weight-decay regime on modular-arithmetic networks.
\end{abstract}

\section{Introduction}

When a neural network keeps training after it has fit its training set, it usually stops improving.
Grokking is the striking exception: on certain algorithmic tasks, test accuracy stays at chance for many
thousands of steps after the training loss has collapsed, then rises sharply to near perfect
\citep{power2022grokking}. The phenomenon is robust and reproducible; the open question is no longer
\emph{whether} it happens but \emph{what controls the delay}.

One variable is repeatedly implicated: the weight norm. Grokking coincides with weight-norm decay
\citep{liu2023omnigrok}, the delay lengthens when the norm is held high, and a recent line treats norm
minimization on the zero-loss manifold as the organizing principle
\citep{musat2025geometry,boursier2025framework}. If the norm is clamped to a chosen value throughout
training, the grokking delay follows a clean exponential dose response in the held norm. It is tempting to
read this as the weight norm setting the timescale of grokking.

We argue that this reading conflates two things the clamp changes together. Holding $\|W\|$ fixed requires
rescaling the weight matrices, and rescaling also raises the logits, which in cross-entropy pushes the
softmax toward saturation. So when a higher held norm lengthens the delay, we cannot yet tell whether the
scalar norm is the operative variable or whether it is the logit scale that the rescaling drags along.
The distinction matters: the first says grokking is governed by a geometric quantity, the second points at
a function-space quantity already implicated by the numerical-stability account of grokking
\citep{prieto2025grokking}.

We separate the two with an intervention. We hold the total weight norm fixed with the clamp and add a
non-trainable output temperature $\tau$ that divides the logits before the loss. Because $\tau$ is not a
weight, it is not part of $\|W\|$; at a clamped norm it tunes the effective logit scale while leaving the
norm fixed. The result, across two moduli, is unambiguous under cross-entropy: varying $\tau$ alone slides
the grokking delay across the entire range produced by raising the norm, and the baseline and the
norm-raised runs lie on a single curve of delay against effective logit scale. Matching the logit scale
back to baseline recovers $0.83$ and $0.89$ of the norm-induced delay. The weight-norm effect on the delay
is, to that extent, the logit-scale effect.

The mechanism is loss-dependent. Under mean-squared error the effective logit scale at grokking is pinned
near one because the regression target fixes it, $\tau$ cannot move it, and the norm effect, while still
present, is about half the size and is not carried by the logit scale. A control rules out the obvious
artifact: $\tau$ leaves the memorization time essentially unchanged and acts almost entirely on the
delayed-generalization phase, so the effect is about grokking specifically and not about training speed.

Our contributions:
\begin{itemize}
\item A temperature-mediation test that separates the scalar weight norm from the effective logit scale at
a fixed clamped norm, and the finding that under cross-entropy the norm-induced grokking delay is recovered
primarily ($\sim$0.85, tight bootstrap CI, two moduli) by restoring the logit scale.
\item A data collapse: across a grid of norms and temperatures, the delay is a function of the effective
logit scale alone ($R^2=0.97$), with the norm dose adding $1$--$2\%$ beyond it.
\item A clean loss dissociation: the logit-scale channel is active under cross-entropy and absent under
mean-squared error, so the weight-norm dependence of grokking is not a single mechanism.
\item A memorization control showing the temperature acts on the delay and not on memorization, a float64
audit that reaches the same softmax-saturation channel independently, and qualitative corroboration in a
no-LayerNorm transformer.
\item A same-state test that forks arms from one identical state and shows the delay tracks the held norm
value, not the clamp's rescaling operation, closing the rescaling-artifact objection for this setting.
\item Honest scope: the result localizes the proximal variable under cross-entropy; it does not
characterize the mean-squared-error route, and the quantitative estimates are for MLPs at two moduli.
\end{itemize}

\section{Setup}

\textbf{Task and model.} We study modular addition, inputs $(a,b)\in\{0,\dots,p-1\}^2$ and target
$(a+b)\bmod p$, a $p$-way classification, for $p\in\{43,59,67,97,113\}$. The dataset is all $p^2$ pairs; a
fixed fraction $\alpha=0.40$ is the training split, drawn per seed by a seeded permutation. The model is a
two-layer MLP: each of $a,b$ is embedded through a shared $E\in\mathbb{R}^{p\times d}$ ($d=128$), the two
embeddings are concatenated and passed through a linear layer $W_1\in\mathbb{R}^{2d\times H}$ ($H=256$)
with a GeLU nonlinearity, then a linear readout $W_2\in\mathbb{R}^{H\times p}$. Optimizer is AdamW
($\beta_1=0.9$, $\beta_2=0.999$, learning rate $10^{-3}$, weight decay $\lambda=1.0$ unless varied),
full-batch unless noted, $12$ seeds per cell. The weight norm is
$\|W\|=\sqrt{\|E\|_F^2+\|W_1\|_F^2+\|W_2\|_F^2}$, biases excluded.

\textbf{The clamp.} After each optimizer step from $t\ge t_{\mathrm{int}}$, the weight matrices are
rescaled by a single scalar so that $\|W\|=\rho\,w_c$ exactly, where $w_c$ is the norm at grokking measured
in a free control run and $\rho$ is the dose. The clamp is engaged before memorization in every cell used
here, so the intervention is active throughout the relevant dynamics.

\textbf{Effective logit scale.} The central variable of this paper needs a precise definition. For a
configuration with temperature $\tau$, the effective logits are the post-$\tau$ logits the loss actually
sees. The effective logit scale is the mean, over all $p^2$ input pairs and over seeds, of the $L_2$ norm
of the per-example effective logit vector, read at the grokking step (the logged step nearest the median
$T_{\mathrm{grok}}$ of that cell). In the mediation analysis it is normalized by the baseline ($\rho=1$,
$\tau=1$) value so the axis is dimensionless and comparable across moduli.

\textbf{Temperatures and metrics.} The temperature $\tau$ divides the logits before the softmax (CE) or the
squared error (MSE); $\tau=1$ recovers the standard model and $\tau$ is held fixed within a run. We log
test accuracy (grokking time $T_{\mathrm{grok}}$ is the first step with test accuracy $\ge 0.90$, median
over seeds), training accuracy (memorization time $T_{\mathrm{mem}}$, $\ge 0.99$), the per-group weight
norms, the effective logit scale, and the softmax-collapse rate of \citet{prieto2025grokking}.

\section{The fixed-norm exponential law (the dose response)}

Holding the norm fixed and sweeping the dose $\rho$ yields a clean dose response between the held norm and
the grokking time. Across all five moduli the relationship is exponential,
$T_{\mathrm{grok}}\propto\exp(\alpha\|W\|)$, with a per-modulus slope $\alpha$. We compare this against a
power law and against a saddle-node form $T\propto(w_0-\|W\|)^{-1/2}$ (the signature a fold or
critical-slowing-down bifurcation would produce) by AIC. The exponential is generally selected
(Table~\ref{tab:a1}): decisively at four of the five moduli ($\Delta$AIC $4.6$--$6.4$), and weakly at
$p=43$ ($\Delta$AIC $0.4$, essentially tied with the power-law), which we report rather than smooth over.
We read this as a robust empirical regularity of the fixed-norm dose response, not a law derived from
first principles, and we use it here only as the readout against which the interventions below are
measured. The slopes are fit on the softmax-collapse-free range ($\rho\le 1.15$; \S\ref{sec:sc}).

This dose response is the phenomenon to be explained. It establishes that raising the held norm lengthens
the delay; it does not, on its own, say what the norm acts through. That is the question of the next
section.

\begin{table}[h]\centering\small
\caption{Fixed-norm exponential law $\ln T_{\mathrm{grok}}=c+\alpha\|W\|$ per modulus, fit on the
softmax-collapse-free range ($\rho\le 1.15$), majority-grok cells only. $\Delta$AIC is the exponential's
margin over the better of the power-law and saddle-node forms (positive favors the exponential).}
\label{tab:a1}
\begin{tabular}{rrrrrc}
\toprule
$p$ & $n$ & $\alpha$ & $R^2$ & $\Delta$AIC & SC-free norm range \\
\midrule
43  & 6 & 0.173 & 0.987 & \textbf{0.4} (weak) & 45--58 \\
59  & 7 & 0.140 & 0.991 & 6.4 & 46--62 \\
67  & 7 & 0.126 & 0.982 & 5.5 & 48--65 \\
97  & 7 & 0.094 & 0.957 & 4.6 & 56--75 \\
113 & 7 & 0.087 & 0.959 & 4.6 & 60--81 \\
\bottomrule
\end{tabular}
\end{table}

\section{What the weight norm acts through}

\subsection{A temperature-mediation test}\label{sec:med}

The clamp holds $\|W\|$ fixed by rescaling the weights, but that rescaling raises the logits, and in
cross-entropy larger logits saturate the softmax. So the dose response of \S3 does not separate the scalar
norm from the logit scale it sets. We separate them with a non-trainable output temperature $\tau$ that
divides the logits before the loss and is not counted in $\|W\|$. At a clamped norm, $\tau$ tunes the
effective logit scale while leaving the norm fixed.

The design is a three-condition mediation. Baseline ($\rho=1.0$, $\tau=1$) gives delay $T_0$; raising the
norm ($\rho=1.15$, $\tau=1$) gives $T_1$; and at $\rho=1.15$ we sweep $\tau$ upward, which lowers the
effective logit scale at fixed norm. If the norm effect runs through the logit scale, setting $\tau$ so
that the effective logit scale matches the baseline should recover the baseline delay.

\begin{figure}[t]\centering
\includegraphics[width=\linewidth]{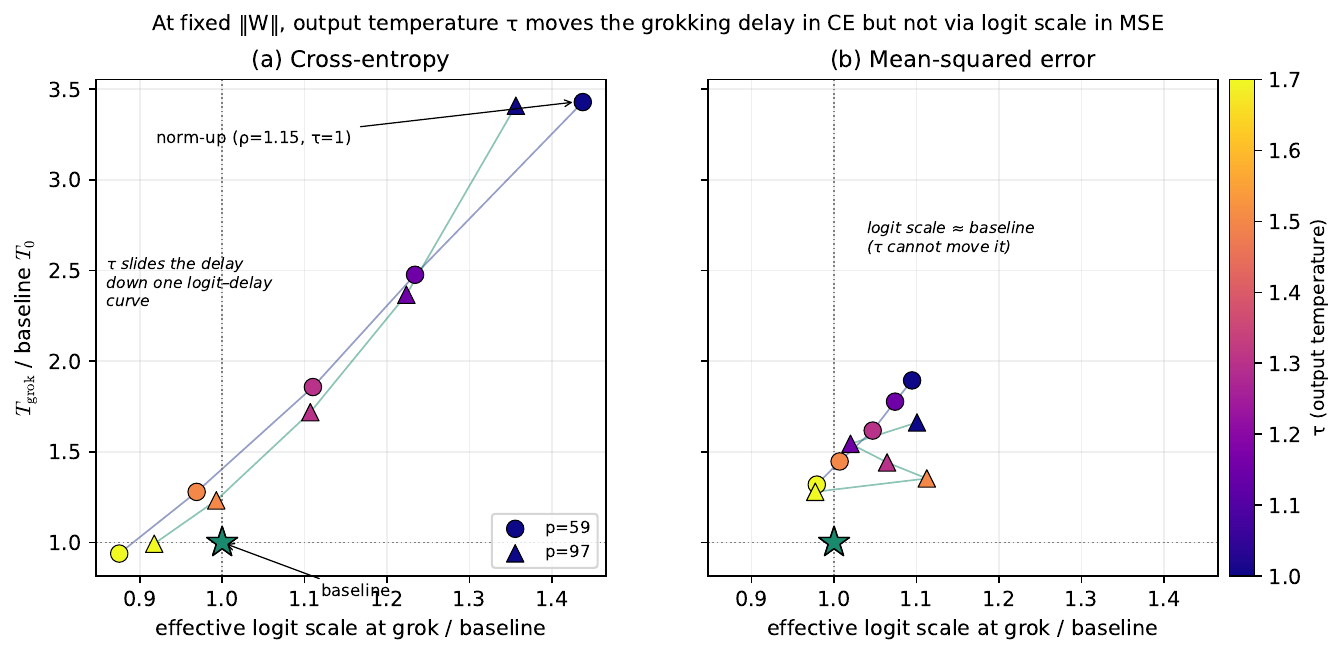}
\caption{At a fixed clamped norm, varying only the output temperature $\tau$. \textbf{(a)} Under
cross-entropy the cells trace one curve of delay against effective logit scale: the baseline (star) and
the norm-up point ($\tau=1$, top right) lie on it, and increasing $\tau$ slides the delay back down toward
baseline. About $83$--$89\%$ of the norm-up delay is recovered by matching the logit scale (two moduli).
\textbf{(b)} Under mean-squared error the effective logit scale at grokking is pinned near baseline and
$\tau$ cannot move it; the residual norm effect (vertical spread) is not along the logit-scale axis. Both
moduli overlaid; axes normalized by the baseline cell.}
\label{fig:med}
\end{figure}

Under cross-entropy this is what happens (Fig.~\ref{fig:med}a, Table~\ref{tab:main}). The norm-up delay is
$3.4\times$ the baseline at both moduli. Sweeping $\tau$ from $1.0$ to $1.7$ slides the delay monotonically
back down, from $3.4\times$ to below baseline, and the cells trace a single curve of delay against
effective logit scale on which the baseline and norm-up points both lie. The effective logit scale itself
moves by about $1.6\times$ across the sweep. Reading off the delay where the effective logit scale equals
its baseline value gives a \emph{logit-scale recovery fraction} $(T_1-T_2)/(T_1-T_0)$ of $0.83$ at $p=59$
(95\% bootstrap CI $[0.82,0.85]$) and $0.89$ at $p=97$ ($[0.88,0.91]$); the CI is a paired resample over
the twelve seeds and reflects seed variability in $T_{\mathrm{grok}}$, with the logit curve held fixed. We
call it a recovery fraction rather than a mediation fraction deliberately: it is the share of the norm-up
delay that the temperature intervention recovers by restoring the effective logit scale, an
intervention-level quantity, not a regression-based natural-indirect-effect estimate. To that extent the
weight-norm effect on the delay is the logit-scale effect: the norm matters because it sets the logit
scale, which sets the softmax saturation. This is the same channel the audit of \S\ref{sec:sc} probes,
reached here by intervention rather than by precision. The next subsection shows the relationship is not
specific to this one norm level: across a grid of norms and temperatures, the delay collapses onto the
effective logit scale.

\begin{table}[h]\centering\small
\caption{Temperature mediation at $\rho=1.15$ (medians over 12 seeds). The logit-scale recovery fraction is
the share of the norm-up delay recovered by matching the effective logit scale back to baseline; it is
reported only for cross-entropy, where the logit scale is a usable knob.}
\label{tab:main}
\begin{tabular}{llrrrc}
\toprule
loss & $p$ & $T_0$ & $T_1$ & norm-up & recovery fraction \\
\midrule
CE  & 59 & 3675 & 12600 & $3.43\times$ & $0.83\ [0.82,0.85]$ \\
CE  & 97 & 1775 & 6050  & $3.41\times$ & $0.89\ [0.88,0.91]$ \\
MSE & 59 & 1175 & 2225  & $1.89\times$ & n/a (channel inactive) \\
MSE & 97 & 850  & 1412  & $1.66\times$ & n/a (channel inactive) \\
\bottomrule
\end{tabular}
\end{table}

\subsection{A data collapse across the norm--temperature grid}\label{sec:collapse}

\begin{figure}[t]\centering
\includegraphics[width=\linewidth]{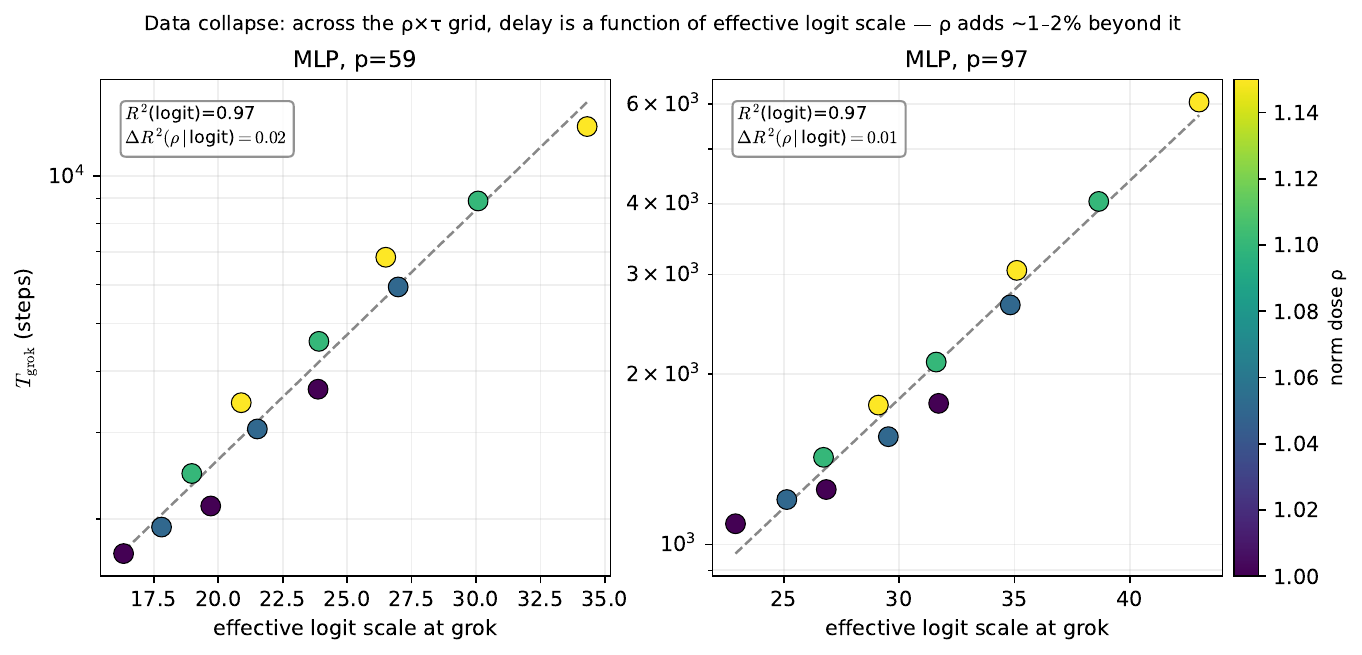}
\caption{Data collapse across the $\rho\times\tau$ grid (12 cells per modulus, all grok 12/12).
$T_{\mathrm{grok}}$ against the effective logit scale at grokking, colored by the norm dose $\rho$. Cells of
different $\rho$ but matched effective logit scale share the same delay, so the delay collapses onto the
logit scale; the norm dose adds $1$--$2\%$ of explained variance beyond it.}
\label{fig:collapse}
\end{figure}

The single-norm recovery fraction asks what happens at one norm level. A stronger question is whether the
delay is a function of the effective logit scale \emph{regardless} of how that scale was reached, whether by
raising the norm or by lowering the temperature. We test this on a grid of four norm doses
($\rho\in\{1.00,1.05,1.10,1.15\}$) crossed with three temperatures ($\tau\in\{1.0,1.3,1.7\}$), twelve cells
per modulus, all grokking 12/12. Figure~\ref{fig:collapse} plots $T_{\mathrm{grok}}$ against the effective
logit scale at grokking, colored by $\rho$: the cells fall on a single log-linear curve, and cells of
different $\rho$ at the same effective logit scale have the same delay.

We quantify the collapse by regression. A least-squares fit of $\ln T_{\mathrm{grok}}$ on the effective
logit scale explains $R^2=0.97$ of the variance at both moduli; adding $\rho$ as a second predictor raises
$R^2$ by only $0.02$ ($p=59$) and $0.01$ ($p=97$), so once the effective logit scale is known the norm dose
carries almost no additional information about the delay (Table~\ref{tab:collapse}). The delay is, to this
precision, a function of the effective logit scale alone. We are careful about the causal reading: the
$x$-axis is the \emph{realized} logit scale at grokking and is therefore endogenous, so the collapse is
strong organizing evidence, not an intervention on its own; the causal weight rests on the
temperature manipulation of \S\ref{sec:med}, which sets the logit scale directly at fixed norm. The two are
complementary: the intervention shows the channel is causal, and the collapse shows it accounts for
essentially all of the norm dependence.

\begin{table}[h]\centering\small
\caption{Collapse regression across the $\rho\times\tau$ grid. $R^2$ of $\ln T_{\mathrm{grok}}$ against the
effective logit scale alone, against $\rho$ alone, and against both; $\Delta R^2(\rho\,|\,\text{logit})$ is
the variance $\rho$ adds beyond the logit scale (near zero indicates collapse).}
\label{tab:collapse}
\begin{tabular}{rccccc}
\toprule
$p$ & $R^2$(logit) & $R^2(\rho)$ & $R^2$(both) & $\Delta R^2(\rho\,|\,\text{logit})$ & verdict \\
\midrule
59 & 0.968 & 0.407 & 0.988 & 0.020 & collapse \\
97 & 0.973 & 0.415 & 0.984 & 0.011 & collapse \\
\bottomrule
\end{tabular}
\end{table}

\subsection{The mechanism is loss-dependent}

Mean-squared error dissociates cleanly (Fig.~\ref{fig:med}b). There the effective logit scale at grokking
sits near one and barely moves with $\tau$, because the regression target fixes the output scale, so the
temperature cannot act on a logit-scale channel: across $\tau\in[1.0,1.7]$ the effective logit scale moves
about $1.1\times$, against $1.6\times$ under cross-entropy. The norm-up effect under MSE is real but about
half the size ($1.7$--$1.9\times$ versus $3.4\times$) and is carried by a route that does not pass through
the logit scale. So the weight-norm dependence of the delay is not one mechanism. Under cross-entropy it
is logit-scale saturation; under mean-squared error it is something else.

We are deliberately conservative about the MSE residual. The temperature still shortens the MSE delay
somewhat, but a temperature that divides the logits also changes the scale of the squared-error target, so
we do not read this residual as a clean second channel. We report MSE only to establish the dissociation
that the logit-scale channel is absent there; we do not assign the residual norm effect to a specific
mechanism. This connects to an asymmetry noted by \citet{prieto2025grokking}, who observed that grokking
under softmax cross-entropy behaves differently from the MSE setting of \citet{liu2023omnigrok}, with the
softmax-collapse rate approaching one under cross-entropy; our intervention gives that asymmetry a
mechanism on the cross-entropy side.

\subsection{The effect is on the delay, not on memorization}

\begin{figure}[t]\centering
\includegraphics[width=\linewidth]{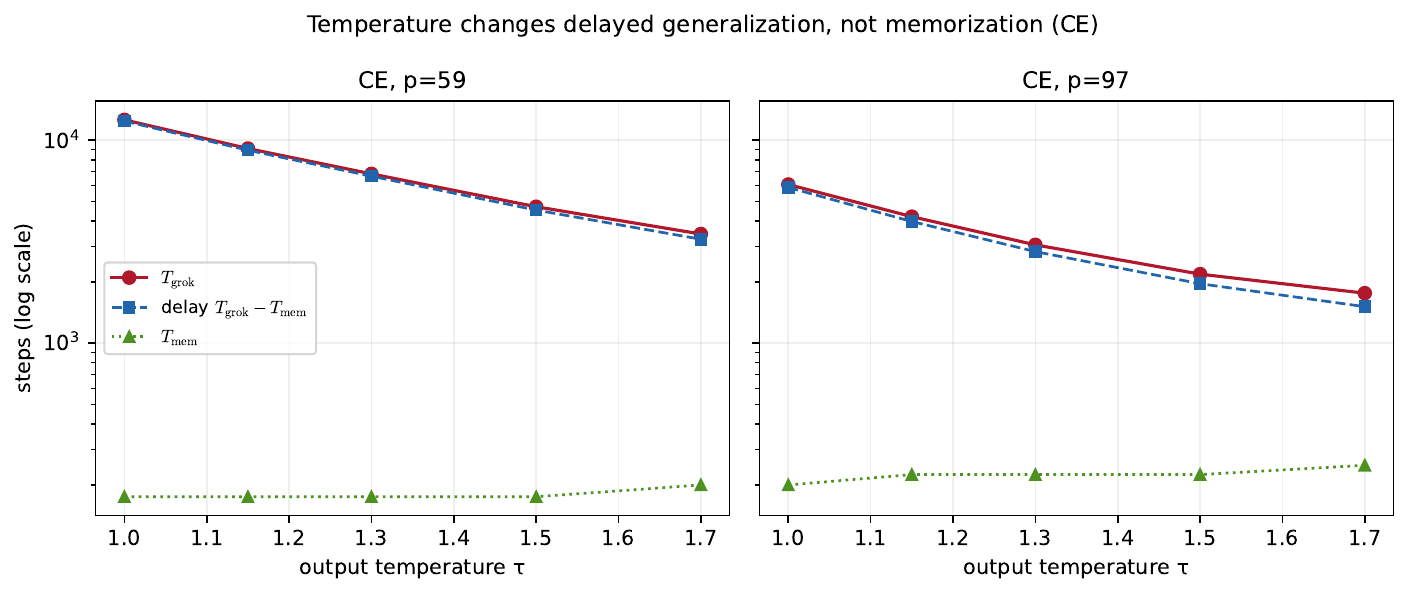}
\caption{Under cross-entropy, the temperature changes the delayed-generalization phase, not memorization.
$T_{\mathrm{mem}}$ (bottom) is flat across $\tau$ while $T_{\mathrm{grok}}$ and the delay
$T_{\mathrm{grok}}-T_{\mathrm{mem}}$ fall together. $y$-axis is log scale.}
\label{fig:mem}
\end{figure}

A temperature that divides the logits changes both the softmax saturation and the gradient magnitude, so a
skeptic can ask whether $\tau$ simply speeds up training in general. It does not, under cross-entropy.
Splitting the per-seed times (Fig.~\ref{fig:mem}), the memorization time $T_{\mathrm{mem}}$ is essentially
flat across $\tau$ ($175\to200$ steps at $p=59$, $200\to250$ at $p=97$), while the delay
$T_{\mathrm{grok}}-T_{\mathrm{mem}}$ carries the entire $\tau$ effect (100\% and 101\% of the change in
$T_{\mathrm{grok}}$). The temperature acts on the delayed-generalization phase, not on how fast the
network fits the training set, so the gradient-magnitude reading does not explain it. (Under MSE the split
is less clean: at $p=97$ about two thirds of the $\tau$ effect falls on $T_{\mathrm{mem}}$, a further
reason we do not interpret the MSE residual.)

\subsection{Corroboration across allocation and architecture}

A second intervention corroborates that the scalar norm is not the sole driver. Holding the total norm
fixed and shifting mass toward the readout layer (raising $\|W_2\|$ at the expense of $\|E\|$) changes the
delay by up to $2.2\times$ at $p=59$ and $1.7\times$ at $p=97$. We treat this as supporting rather than
primary, because reallocating mass also changes embedding capacity, so it is not a clean logit-scale move;
the temperature test is the clean manipulation.

The mechanism also reproduces in a different architecture. We ran the temperature test on a no-LayerNorm
transformer at $p=59$: raising the held norm lengthens the delay, and at the raised norm, increasing the
temperature shortens the delay monotonically by $4.7\times$ ($16{,}100\to3{,}500$ steps), confirming that a
variable outside the norm controls the delay under cross-entropy in a transformer as it does in the MLP. We
report this as qualitative corroboration and deliberately do not quote a recovery fraction for the
transformer: its logits grow far larger and more erratically than the MLP's. The highest-stress cell
shows the uncontrolled logit growth of \citet{prieto2025grokking}, with the logit scale climbing past
$1000$ after grokking, so the effective logit scale at grokking is a noisier proxy there, and matching it
back to baseline would require extrapolation. The direction and size of the effect transfer; the precise
mediation estimate is MLP-specific. That the transformer's logits are harder to hold down is itself
consistent with the saturation account.

We are also explicit about what the temperature test does not do on its own. It shows that, at a fixed
clamped norm, a variable outside the norm controls the delay, which localizes the proximal variable under
cross-entropy to the logit scale, and we claim the logit scale is the proximal mediator of the norm effect.
The remaining question is whether the clamp's rescaling operation, rather than the held value, drives the
delay; the next subsection settles it.

\subsection{A same-state test of the clamp}\label{sec:samestate}

The clamp sets the norm by rescaling the weights, so one objection survives the temperature test: perhaps
the rescaling operation, applied every step, distorts the trajectory, and the delay reflects the operation
rather than the held value. We separate the two by forking four arms from one identical state. We train
freely to a fork step after memorization and before grokking, then continue from the saved state under four
conditions at $\rho=1.15$: free (no clamp), clamp at the forked norm $N_0$ (the operation is active but the
value is unchanged), clamp at $\rho N_0$, and clamp at $N_0/\rho$.

\begin{figure}[t]\centering
\includegraphics[width=\linewidth]{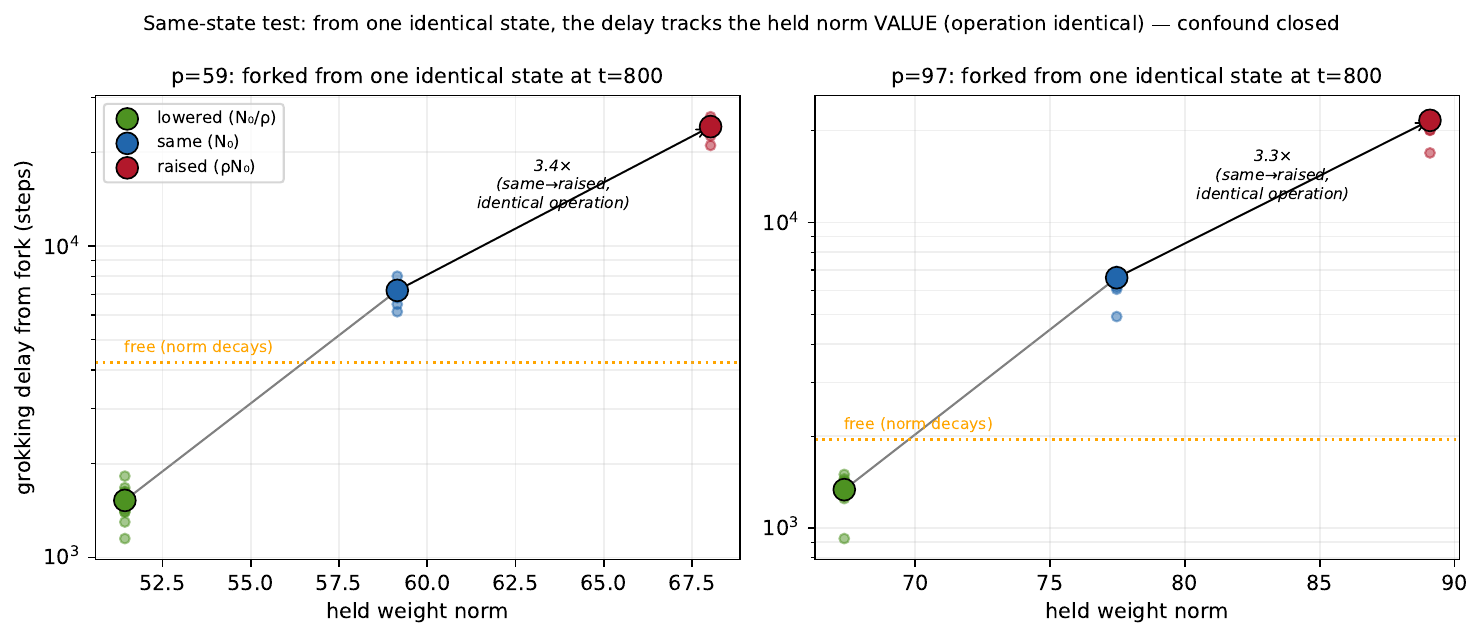}
\caption{Four arms forked from one identical post-memorization state ($t=800$). The grokking delay (from
the fork) rises monotonically with the held norm value across all twelve seeds. The labelled contrast,
clamp-at-$N_0$ versus clamp-at-$\rho N_0$, applies the identical clamp operation and differs only in the
held value ($3.3$--$3.4\times$), so the delay tracks the value, not the rescaling operation. Free (dotted)
groks faster because its norm decays after the fork.}
\label{fig:samestate}
\end{figure}

From the identical state, the delay rises monotonically with the held norm value, in all twelve seeds at
both moduli (Fig.~\ref{fig:samestate}). The decisive contrast is clamp-at-$N_0$ against clamp-at-$\rho N_0$.
Both apply the same clamp operation from the same state and differ only in the held value, and the raised
arm takes $3.36\times$ longer at $p=59$ (95\% CI $[3.27,3.43]$) and $3.27\times$ at $p=97$ ($[3.21,3.38]$).
Since the operation is identical in the two arms, the difference is not an artifact of the operation: the
delay follows the held value. The ordering across lower, hold, and raise also rules out a one-time
rescaling jump, which would perturb the up and down arms the same way instead of ordering them by value.
The free arm groks faster than clamp-at-$N_0$ because its norm decays after the fork, which is the norm
decay that the generalizing-shell account associates with grokking \citep{liu2023omnigrok}. This closes the
rescaling-artifact objection for the cross-entropy MLP setting. Training full-batch leaves the dose
response intact, so the effect is not a sampling-noise phenomenon either.

\section{A softmax-collapse audit}\label{sec:sc}

Because the cross-entropy clamp operates at large logits, the regime in which softmax collapse can occur
\citep{prieto2025grokking}, we audit the law for floating-point artifacts. We re-run the two highest-dose
cross-entropy cells at $p=59$ in float64, holding the norm at the identical absolute values used in
float32, and record the softmax-collapse rate (Table~\ref{tab:sc}). At $\rho=1.15$ the collapse rate is
zero in both precisions and the grokking times agree, so the law is precision-robust there. At $\rho=1.25$
the float32 run shows a 31\% collapse rate and groks at 25k steps, whereas the float64 run does not grok
within 60k, so the float32 collapse accelerates a spurious transition at extreme norm. We therefore fit the
law on the collapse-free range and exclude the affected cell. This audit is convergent with \S\ref{sec:med}:
softmax collapse is the extreme of the same logit-saturation channel that the temperature test isolates,
reached by precision without any intervention.

\begin{table}[h]\centering\small
\caption{Softmax-collapse audit at $p=59$: the two highest-dose cross-entropy cells re-run in float64 at
held norms identical to float32. $T$ = median grok step; sc = softmax-collapse rate.}
\label{tab:sc}
\begin{tabular}{ccccccl}
\toprule
$\rho$ & held $\|W\|$ & $T_{32}$ & $T_{64}$ & sc$_{32}$ & sc$_{64}$ & verdict \\
\midrule
1.15 & 62 & 12.75k & 12.16k & 0.00 & 0.00 & precision-robust \\
1.25 & 68 & 25.4k & not reached by 60k & 0.31 & 0.00 & float32 SC-confounded \\
\bottomrule
\end{tabular}
\end{table}

\section{Related work}

\textbf{Weight norm and norm minimization.} The weight norm has long been central to grokking: Omnigrok
ties grokking to the norm decaying into a generalizing shell \citep{liu2023omnigrok}, and a recent line
characterizes the late dynamics as norm minimization on the zero-loss manifold
\citep{musat2025geometry,boursier2025framework}, including beyond the Euclidean norm
\citep{notsawo2025beyond}. These accounts concern the \emph{endpoint} (which configuration the optimizer
selects once training loss is zero). Our question is about the \emph{timescale} and the proximal variable
that sets it; we find that under cross-entropy the norm acts on the timescale through the logit scale, so
the two are complementary rather than competing.

\textbf{Logit growth, softmax saturation, and optimization starvation.} The account closest to our result
is the numerical-stability view of \citet{prieto2025grokking}: cross-entropy training without
regularization drives uncontrolled logit growth (``naive loss minimization''), whose extreme is softmax
collapse, where floating-point errors zero the gradient and stall learning. Our temperature test isolates
the upstream of that channel, the logit scale, and shows it mediates most of the norm-induced delay under
cross-entropy, with the collapse of \citet{prieto2025grokking} as its endpoint (\S\ref{sec:sc}).
Independently, \citet{beck2026logit} show that regularizing logits directly biases linear classifiers
toward clustered logits and can induce grokking, further evidence that logit space, not weight norm alone,
is the operative arena under cross-entropy. The intervention itself, dividing logits by a fixed
temperature, is the temperature scaling of \citet{guo2017calibration}; we use it not for calibration but as
a handle on the effective logit scale at fixed norm.

\textbf{Phase transitions and scaling.} A parallel line characterizes grokking as a phase transition along
model or data axes: a first-order transition in two-layer networks \citep{rubin2024firstorder}, an
information-theoretic transition \citep{clauw2024infotheoretic}, Ising-style and local-rule descriptions
\citep{hutchison2025ising,zunkovic2024grokking}, and finite-size or dimensional scaling
\citep{bi2026finitesize,wang2026dimensional,wang2026criticality}. These concern the nature of the
transition and its scaling with model or data size, a different axis from our intervention-level question
of which variable gates the delay at fixed data and model.

\textbf{Other proposed drivers.} Grokking has also been attributed to noise-assisted escape and slingshot
dynamics \citep{thilak2022slingshot,lopatin2025predator}, to circuit-efficiency reallocation
\citep{nanda2023progress,varma2023circuit}, and to a lazy-to-rich transition
\citep{kumar2024grokking,lyu2020margin}. Our full-batch result places the delay outside the
noise-activated family for this setting, and our structure measure is the Fourier progress measure of
\citet{nanda2023progress}.

Most of this work asks what \emph{produces} grokking and answers with a mechanism that suffices. Our
question is narrower: with the weight norm held at a chosen value, which mechanism still moves the delay?
The temperature handle lets us put one candidate, the logit scale, in and take it out directly, so the
contribution here is not another mechanism but a way to decide, under cross-entropy, what the norm acts
through.

\section{Limitations}

The claims are confined to the $\ell_2$/weight-decay regime on modular-arithmetic networks. The
quantitative results (the recovery fraction and the data collapse) are established on MLPs at two moduli;
a no-LayerNorm transformer reproduces the effect in direction and size but with noisier logit dynamics, so
we do not extend the precise recovery estimate to it, and we have not tested larger-scale settings or other
task families (sparse parity, group tasks beyond modular arithmetic). The same-state test (\S\ref{sec:samestate})
addresses the clamp-rescaling confound for the cross-entropy MLP setting, but the matched contrast is at one
fork point and one $\rho$; we have not mapped the full fork-time dependence. The data collapse uses the
realized logit scale at grokking, which is endogenous, so it is organizing evidence and the causal weight
rests on the temperature intervention. The mean-squared-error route is identified only negatively: it does
not pass through the logit scale, and we do not characterize it further; the residual temperature effect
under MSE is partly a memorization-speed effect and is not interpreted. The recovery-fraction CI reflects
seed variability in $T_{\mathrm{grok}}$ with the logit curve held fixed, and does not include interpolation
uncertainty on the logit axis. Two directions follow directly: characterizing the mean-squared-error route,
which is left here as not-the-logit-scale, and testing whether the logit-scale mediation holds at
transformer scale, where the logit dynamics we saw are already noisier.

\section{Conclusion}

The weight-norm dependence of the grokking delay is not, under cross-entropy, an effect of the scalar norm.
Holding the norm fixed and varying only an output temperature slides the delay across the full norm-up
range, and matching the effective logit scale back to baseline recovers about $85\%$ of the delay at two
moduli, with the effect falling on the delayed-generalization phase and not on memorization. Across a
grid of norms and temperatures the delay collapses onto the effective logit scale, which alone explains
$97\%$ of its variance. The weight norm is an upstream handle; the proximal variable is the logit scale,
and the channel is the softmax saturation that the numerical-stability account of grokking also points to.
The mechanism is loss-dependent: mean-squared error pins the logit scale and the norm effect runs
elsewhere. Forking arms from one identical state confirms that the delay follows the held norm value and
not the rescaling operation, so the dose response is not a clamp artifact. Beyond grokking, the
methodological point is that a causal claim about a scalar quantity such as the weight norm has to be
separated from the function-space variables that rescaling the weights changes along with it; here, once
that separation is made, much of what looked like a weight-norm law is a logit-scale law.

\appendix
\bigskip
\begin{center}\large\textbf{Appendix}\end{center}

\section{Model, task, and exact definitions}

\textbf{Task.} Modular addition: inputs $(a,b)\in\{0,\dots,p-1\}^2$, target $(a+b)\bmod p$, a $p$-way
classification, for $p\in\{43,59,67,97,113\}$ (the dose-response law) and $p\in\{59,97\}$ (the
interventions). The dataset is all $p^2$ pairs; a fixed fraction $\alpha=0.40$ is the training split,
drawn per seed by a seeded permutation, the remainder held out for test.

\textbf{Model.} A two-layer MLP. Each of $a,b$ is mapped through a shared embedding
$E\in\mathbb{R}^{p\times d}$ ($d=128$); the two embeddings are concatenated and passed through a linear
layer $W_1\in\mathbb{R}^{2d\times H}$ ($H=256$) with bias $b_1$ and a GeLU nonlinearity, then a linear
readout $W_2\in\mathbb{R}^{H\times p}$ with bias $b_2$. Weight matrices are initialized
$\mathcal{N}(0,1)/\sqrt{\mathrm{fan\text{-}in}}$; biases initialized to zero.

\textbf{Optimizer.} AdamW ($\beta_1=0.9$, $\beta_2=0.999$, $\epsilon=10^{-8}$), learning rate $10^{-3}$,
weight decay $\lambda=1.0$ unless varied. Full-batch gradients unless a minibatch noise axis is specified.
Weight decay and the clamp act on $E,W_1,W_2$; biases are unregularized and excluded from $\|W\|$.
Twelve seeds per cell.

\textbf{Weight norm.} $\|W\|=\sqrt{\|E\|_F^2+\|W_1\|_F^2+\|W_2\|_F^2}$. This is the quantity the clamp
holds and the $x$-axis of the dose response.

\textbf{The clamp.} After each optimizer step with $t\ge t_{\mathrm{int}}$ ($t_{\mathrm{int}}=500$ for the
cross-entropy runs, $50$ for the mean-squared-error runs, chosen below memorization so the intervention
is active throughout the relevant dynamics), the weight matrices are rescaled by a single scalar so that
$\|W\|=\rho\,w_c$ exactly, where $w_c$ is the median norm at grokking in a free control run and $\rho$ is
the dose. The control calibrates $w_c=54.49$ ($p=59$) and $65.85$ ($p=97$) under cross-entropy.

\textbf{Temperature.} A non-trainable scalar $\tau$ divides the logits before the loss; $\tau=1$ recovers
the standard model. $\tau$ is not a parameter and is not part of $\|W\|$, so at a clamped norm it changes
the effective logit scale while leaving $\|W\|$ fixed.

\textbf{Effective logit scale.} For a configuration with temperature $\tau$, the effective logits are the
post-$\tau$ logits the loss sees. The effective logit scale is
$\frac{1}{S}\sum_s \frac{1}{p^2}\sum_{(a,b)} \|z_{s,(a,b)}/\tau\|_2$, the mean over all $p^2$ input pairs
and over the $S=12$ seeds of the $L_2$ norm of the per-example effective logit vector $z/\tau$, read at
the grokking step (the logged step nearest the median $T_{\mathrm{grok}}$ of the cell). In the mediation
analysis it is normalized by the baseline ($\rho=1$, $\tau=1$) value.

\textbf{Times.} $T_{\mathrm{mem}}$ is the first step with train accuracy $\ge 0.99$; $T_{\mathrm{grok}}$ is
the first step with test accuracy $\ge 0.90$. Both are per seed and aggregated by the median over seeds. A
cell is counted only if a majority of seeds grok.

\section{Model selection for the dose-response law}

For a dose response of $n$ cells with held norms $\|W\|_i$ and log grokking times $\ell_i=\ln T_i$, each
candidate form is fit by least squares and scored by $\mathrm{AIC}=n\ln(\mathrm{RSS}/n)+2k$ with $k$ the
parameter count. The compared forms are the exponential $\ell=c+\alpha\|W\|$ ($k=2$), a power law with free
offset $\ell=c+\beta\ln(\|W\|-w_0)$ ($k=3$), and a saddle-node form $\ell=c-\tfrac12\ln(w_0-\|W\|)$,
$w_0>\max\|W\|$ ($k=2$), the critical-slowing-down signature a fold bifurcation would produce. Only
majority-grokked cells on the softmax-collapse-free range ($\rho\le 1.15$) enter the fit. $\Delta$AIC in
Table~\ref{tab:a1} is the exponential's margin over the better of the power-law and saddle-node forms;
positive favors the exponential.

\section{Mediation: per-cell data and the recovery fraction}

Tables~\ref{tab:medce}--\ref{tab:medmse} give every cell behind \S\ref{sec:med} (medians over 12 seeds;
all cells grok 12/12). Each block holds the baseline ($\rho=1.0$, $\tau=1$) and the norm-up sweep
($\rho=1.15$, $\tau$ increasing, which lowers the effective logit scale at fixed norm). The logit-scale
recovery fraction is $(T_1-T_2)/(T_1-T_0)$, where $T_0$ is the baseline delay, $T_1$ the norm-up delay at
$\tau=1$, and $T_2$ the delay read where the effective logit scale, interpolated over the $\tau$ sweep,
equals the baseline value. The 95\% CI is a paired bootstrap over the twelve seeds (4000 resamples): each
resample recomputes the per-cell median $T_{\mathrm{grok}}$ and the interpolated $T_2$, with the
effective-logit curve held at its full-sample value; the interval is the 2.5--97.5 percentile range. It
therefore reflects seed variability in $T_{\mathrm{grok}}$, the dominant source, and not interpolation
uncertainty on the logit axis.

\begin{table}[h]\centering\small
\caption{Temperature sweep, cross-entropy. $L$ = effective logit scale at grok. All cells grok 12/12.}
\label{tab:medce}
\begin{tabular}{llrrrr}
\toprule
$p$ & cell & $T_{\mathrm{grok}}$ & $T_{\mathrm{mem}}$ & delay & $L$ \\
\midrule
59 & $\rho{=}1.0,\tau{=}1$ (baseline) & 3675 & 175 & 3500 & 23.87 \\
59 & $\rho{=}1.15,\tau{=}1.00$ & 12600 & 175 & 12425 & 34.31 \\
59 & $\rho{=}1.15,\tau{=}1.15$ & 9100 & 175 & 8925 & 29.46 \\
59 & $\rho{=}1.15,\tau{=}1.30$ & 6825 & 175 & 6650 & 26.50 \\
59 & $\rho{=}1.15,\tau{=}1.50$ & 4700 & 175 & 4525 & 23.13 \\
59 & $\rho{=}1.15,\tau{=}1.70$ & 3450 & 200 & 3250 & 20.89 \\
\midrule
97 & $\rho{=}1.0,\tau{=}1$ (baseline) & 1775 & 200 & 1575 & 31.71 \\
97 & $\rho{=}1.15,\tau{=}1.00$ & 6050 & 200 & 5850 & 43.01 \\
97 & $\rho{=}1.15,\tau{=}1.15$ & 4200 & 225 & 3975 & 38.80 \\
97 & $\rho{=}1.15,\tau{=}1.30$ & 3050 & 225 & 2825 & 35.10 \\
97 & $\rho{=}1.15,\tau{=}1.50$ & 2188 & 225 & 1962 & 31.49 \\
97 & $\rho{=}1.15,\tau{=}1.70$ & 1762 & 250 & 1512 & 29.10 \\
\bottomrule
\end{tabular}
\end{table}

\begin{table}[h]\centering\small
\caption{Temperature sweep, mean-squared error. The effective logit scale $L$ is pinned near baseline and
does not move with $\tau$, so no recovery fraction is defined. All cells grok 12/12.}
\label{tab:medmse}
\begin{tabular}{llrrrr}
\toprule
$p$ & cell & $T_{\mathrm{grok}}$ & $T_{\mathrm{mem}}$ & delay & $L$ \\
\midrule
59 & $\rho{=}1.0,\tau{=}1$ (baseline) & 1175 & 250 & 925 & 0.95 \\
59 & $\rho{=}1.15,\tau{=}1.00$ & 2225 & 350 & 1875 & 1.04 \\
59 & $\rho{=}1.15,\tau{=}1.15$ & 2088 & 338 & 1750 & 1.02 \\
59 & $\rho{=}1.15,\tau{=}1.30$ & 1900 & 325 & 1575 & 0.99 \\
59 & $\rho{=}1.15,\tau{=}1.50$ & 1700 & 325 & 1375 & 0.95 \\
59 & $\rho{=}1.15,\tau{=}1.70$ & 1550 & 325 & 1225 & 0.93 \\
\midrule
97 & $\rho{=}1.0,\tau{=}1$ (baseline) & 850 & 600 & 250 & 0.59 \\
97 & $\rho{=}1.15,\tau{=}1.00$ & 1412 & 975 & 438 & 0.65 \\
97 & $\rho{=}1.15,\tau{=}1.15$ & 1312 & 888 & 425 & 0.60 \\
97 & $\rho{=}1.15,\tau{=}1.30$ & 1225 & 825 & 400 & 0.62 \\
97 & $\rho{=}1.15,\tau{=}1.50$ & 1150 & 775 & 375 & 0.65 \\
97 & $\rho{=}1.15,\tau{=}1.70$ & 1088 & 750 & 338 & 0.57 \\
\bottomrule
\end{tabular}
\end{table}

Under cross-entropy the recovery fraction is $0.83$ at $p=59$ (CI $[0.82,0.85]$) and $0.89$ at $p=97$
(CI $[0.88,0.91]$). The memorization control reads directly from these tables: $T_{\mathrm{mem}}$ moves by
$25$--$50$ steps across the full $\tau$ sweep while the delay moves by thousands, so the entire $\tau$
effect is on the delay. Under mean-squared error the effective logit scale moves only $\sim 1.1\times$
across $\tau$ (against $\sim 1.6\times$ under cross-entropy), so no logit-scale channel is available for
$\tau$ to act on; at $p=97$ roughly two thirds of the small $\tau$ effect falls on $T_{\mathrm{mem}}$,
which is a further reason the mean-squared-error residual is not interpreted.

\textbf{Collapse grid.} Table~\ref{tab:gridcells} lists the twelve $\rho\times\tau$ cells per modulus
behind the collapse of \S\ref{sec:collapse} (all grok 12/12). The regression statistics are in
Table~\ref{tab:collapse}.

\begin{table}[h]\centering\small
\caption{The $\rho\times\tau$ grid behind Fig.~\ref{fig:collapse}: $T_{\mathrm{grok}}$ (median over 12
seeds) and the effective logit scale $L$ at grok, cross-entropy. Cells of different $\rho$ but matched $L$
share the delay.}
\label{tab:gridcells}
\begin{tabular}{rrrrrr}
\toprule
& \multicolumn{2}{c}{$p=59$} & & \multicolumn{2}{c}{$p=97$} \\
\cmidrule(lr){2-3}\cmidrule(lr){5-6}
$(\rho,\tau)$ & $T_{\mathrm{grok}}$ & $L$ & & $T_{\mathrm{grok}}$ & $L$ \\
\midrule
$(1.00,1.0)$ & 3675 & 23.9 & & 1775 & 31.7 \\
$(1.00,1.3)$ & 2125 & 19.6 & & 1250 & 26.7 \\
$(1.00,1.7)$ & 1700 & 16.4 & & 1088 & 22.9 \\
$(1.05,1.0)$ & 5938 & 24.9 & & 2650 & 35.0 \\
$(1.05,1.3)$ & 3050 & 20.9 & & 1550 & 29.4 \\
$(1.05,1.7)$ & 1925 & 17.9 & & 1200 & 26.7 \\
$(1.10,1.0)$ & 8888 & 29.3 & & 4038 & 38.7 \\
$(1.10,1.3)$ & 4600 & 23.2 & & 2100 & 31.6 \\
$(1.10,1.7)$ & 2475 & 19.6 & & 1425 & 26.8 \\
$(1.15,1.0)$ & 12600 & 34.3 & & 6050 & 43.0 \\
$(1.15,1.3)$ & 6825 & 26.5 & & 3050 & 35.1 \\
$(1.15,1.7)$ & 3450 & 20.9 & & 1762 & 29.1 \\
\bottomrule
\end{tabular}
\end{table}

\textbf{No-LayerNorm transformer.} Table~\ref{tab:tf} gives the transformer temperature sweep at $p=59$
(all grok 12/12). At the raised norm, increasing $\tau$ shortens the delay monotonically by $4.7\times$.
The effective logit scale at grok is large and non-monotonic in $\tau$ (the highest-stress $\tau=1.0$ cell
exhibits post-grokking logit growth past $1000$), so we report this as qualitative corroboration and do not
compute a recovery fraction for the transformer.

\begin{table}[h]\centering\small
\caption{No-LayerNorm transformer, cross-entropy, $p=59$ (medians over 12 seeds). The delay responds to
the held norm and to $\tau$ as in the MLP; the effective logit scale $L$ is larger and noisier.}
\label{tab:tf}
\begin{tabular}{lrrr}
\toprule
cell & $T_{\mathrm{grok}}$ & $L$ at grok & grok \\
\midrule
$\rho{=}1.0,\tau{=}1$ (baseline) & 1238 & 208 & 12/12 \\
$\rho{=}1.15,\tau{=}1.00$ & 16112 & 165 & 12/12 \\
$\rho{=}1.15,\tau{=}1.15$ & 9525 & 153 & 12/12 \\
$\rho{=}1.15,\tau{=}1.30$ & 6525 & 138 & 12/12 \\
$\rho{=}1.15,\tau{=}1.50$ & 4712 & 164 & 12/12 \\
$\rho{=}1.15,\tau{=}1.70$ & 3462 & 177 & 12/12 \\
\bottomrule
\end{tabular}
\end{table}

\section{Softmax-collapse audit and the layer-allocation control}

\textbf{Softmax-collapse audit.} The audit of \S\ref{sec:sc} re-runs the two highest-dose cross-entropy
cells at $p=59$ in float64, at held norms identical to the float32 run, and records the softmax-collapse
rate (the fraction of training points whose correct-class softmax probability reaches one in working
precision). Details in Table~\ref{tab:sc}.

\textbf{Layer allocation (Arm A).} As a second intervention we hold the total norm fixed at $\rho=1.0$
($\|W\|=w_c$) and shift mass between the embedding and the readout by a parameter $\gamma$: the readout
share of the $E$--$W_2$ power budget is scaled by $(1+\gamma)$ (clipped), with $W_1$ held at its baseline
share, so $\gamma>0$ moves mass into $W_2$ at fixed total norm. Table~\ref{tab:alloc} reports the result.
The grokking time changes by $2.2\times$ ($p=59$) and $1.7\times$ ($p=97$) across the sweep at fixed total
norm, which corroborates that the scalar norm is not the sole driver. We treat it as supporting, not
primary for two reasons: the effective logit scale at grok barely moves across $\gamma$ (it is the
\emph{total} readout magnitude, not its share, that the temperature test varies cleanly), so the change is
driven by the embedding-capacity shift rather than by a clean logit-scale move; and at $\gamma=+0.6$,
$p=59$, three of twelve seeds fail to grok within budget. The temperature test of \S\ref{sec:med} is the
clean manipulation.

\begin{table}[h]\centering\small
\caption{Layer allocation at fixed total norm ($\rho=1.0$). $\gamma>0$ shifts mass into the readout
$W_2$. $L$ = effective logit scale at grok. Medians over 12 seeds unless the grok count notes otherwise.}
\label{tab:alloc}
\begin{tabular}{rrrrrrc}
\toprule
$p$ & $\gamma$ & $T_{\mathrm{grok}}$ & $\|W\|$ & $\|E\|$ & $\|W_2\|$ & $L$ \\
\midrule
59 & $-0.60$ & 2975 & 54.5 & 28.4 & 15.6 & 20.60 \\
59 & $-0.30$ & 4012 & 54.5 & 25.0 & 20.6 & 23.35 \\
59 & $+0.00$ & 3962 & 54.5 & 21.0 & 24.7 & 24.46 \\
59 & $+0.30$ & 2862 & 54.5 & 16.1 & 28.1 & 24.01 \\
59 & $+0.60$ & 1825$^{\dagger}$ & 54.5 & 8.8 & 31.2 & 24.84 \\
\midrule
97 & $-0.60$ & 1462 & 65.9 & 39.0 & 21.1 & 27.35 \\
97 & $-0.30$ & 1950 & 65.9 & 34.5 & 27.9 & 31.34 \\
97 & $+0.00$ & 1975 & 65.9 & 29.3 & 33.3 & 32.43 \\
97 & $+0.30$ & 1538 & 65.9 & 22.9 & 38.0 & 31.91 \\
97 & $+0.60$ & 1150 & 65.9 & 13.8 & 42.1 & 32.62 \\
\bottomrule
\end{tabular}
\\[2pt]
{\footnotesize $^{\dagger}$ 9 of 12 seeds grok within budget; the median is over the grokked seeds.}
\end{table}

\section{Same-state arms}

Table~\ref{tab:samestate} gives the four arms forked from one identical state at $t=800$ (\S\ref{sec:samestate}),
all grokking 12/12. The delay is measured from the fork. The recovery contrast is clamp-at-$N_0$ against
clamp-at-$\rho N_0$, both applying the same clamp operation from the same state.

\begin{table}[h]\centering\small
\caption{Same-state arms, cross-entropy, fork at $t=800$ (median delay from fork over 12 seeds). The
raised/same ratio is a paired bootstrap over the 12 seeds. The monotonic ordering holds in all 12 seeds at
both moduli.}
\label{tab:samestate}
\begin{tabular}{lrrr}
\toprule
arm (held norm) & delay $p=59$ & delay $p=97$ \\
\midrule
clamp-lowered ($N_0/\rho$) & 1525 & 1338 \\
free (norm decays) & 4225 & 1950 \\
clamp-same ($N_0$) & 7200 & 6600 \\
clamp-raised ($\rho N_0$) & 24162 & 21600 \\
\midrule
raised/same ratio & $3.36\ [3.27,3.43]$ & $3.27\ [3.21,3.38]$ \\
\bottomrule
\end{tabular}
\end{table}

\section*{Code and data availability}
The clamp and temperature-mediation runner, the per-cell metric outputs used for every fit, table, and
figure, the AIC model-selection script, and the float64 softmax-collapse audit are available at
\url{https://github.com/ClevixLab/grokking-logit-scale}, with seeds fixed so that every number in Tables~\ref{tab:a1}--\ref{tab:sc} and the
figures can be regenerated.

\bibliographystyle{plainnat}
\bibliography{refs}
\end{document}